\documentclass[journal]{IEEEtran}
\usepackage{amsmath}
\usepackage{amssymb}
\usepackage{booktabs}
\usepackage{multirow}
\usepackage{graphicx}
\usepackage{subfigure}
\usepackage{cite}
\usepackage[ruled]{algorithm2e}
\usepackage{array}
\usepackage{color}
\usepackage [latin1]{inputenc}

\begin{document}
\title{Super-Resolution of Brain MRI Images using Overcomplete Dictionaries and Nonlocal Similarity }
%
%
% author names and IEEE memberships
% note positions of commas and nonbreaking spaces ( ~ ) LaTeX will not break
% a structure at a ~ so this keeps an author's name from being broken across
% two lines.
% use \thanks{} to gain access to the first footnote area
% a separate \thanks must be used for each paragraph as LaTeX2e's \thanks
% was not built to handle multiple paragraphs
%
\author{Yinghua~Li,
        Bin~Song$^*$,~\IEEEmembership{Member,~IEEE,}
        Jie~Guo,~\IEEEmembership{Member,~IEEE,}
        Xiaojiang~Du,~\IEEEmembership{Senior Member,~IEEE,}
        Mohsen~Guizani,~\IEEEmembership{Fellow,~IEEE}

\thanks{This work was supported by the National Natural Science Foundation of China under Grant (Nos. 61772387 and 61802296), the Fundamental Research Funds for the Central Universities (JB180101), China Postdoctoral Science Foundation Grant (No. 2017M620438), Fundamental Research Funds of Ministry of Education and China Mobile (MCM20170202), and also supported by the ISN State Key Laboratory.}
\thanks{Y. Li, B. Song and J. Guo are with the State Key Laboratory of Integrated Services Networks, Xidian University, 710071, China. Bin Song is the corresponding author. Emails: liyh@stu.xidian.edu.cn(Y. Li), bsong@mail.xidian.edu.cn(B. Song), jguo@xidian.edu.cn(J. Guo).

 X. Du is with Dept. of Computer and Information Sciences, Temple University, Philadelphia PA, 19122, USA (email: dxj@ieee.org)

 M. Guizani is with Dept. of College of Engineering, Qatar University, Qatar (email: mguizani@gmail.com)}}% <-this % stops a space

%\thanks{Manuscript received April 19, 2005; revised August 26, 2015.}}
% The paper headers
%\markboth{Journal of \LaTeX\ Class Files,~Vol.~14, No.~8, August~2015}%
%{Shell \MakeLowercase{\textit{et al.}}: Bare Demo of IEEEtran.cls for IEEE Journals}
% If you want to put a publisher's ID mark on the page you can do it like
% this:
%\IEEEpubid{0000--0000/00\$00.00~\copyright~2015 IEEE}
% Remember, if you use this you must call \IEEEpubidadjcol in the second
% column for its text to clear the IEEEpubid mark.
% use for special paper notices
%\IEEEspecialpapernotice{(Invited Paper)}
% make the title area
\maketitle
\begin{abstract}
Recently, the Magnetic Resonance Imaging (MRI) images have limited and unsatisfactory resolutions due to various constraints such as physical, technological and economic considerations. Super-resolution techniques can obtain high-resolution MRI images. The traditional methods obtained the resolution enhancement of brain MRI by interpolations, affecting the accuracy of the following diagnose process. The requirement for brain image quality is fast increasing.  In this paper, we propose an image super-resolution (SR) method based on overcomplete dictionaries and inherent similarity of an image to recover the high-resolution (HR) image from a single low-resolution (LR) image. We use the linear relationship among images in the measurement domain and frequency domain to classify image blocks into smooth, texture and edge feature blocks in the measurement domain. The dictionaries for different blocks are trained using different categories. Consequently, an LR image block of interest may be reconstructed using the most appropriate dictionary. Additionally, we explore the nonlocal similarity of the image to tentatively search for similar blocks in the whole image and present a joint reconstruction method based on compressive sensing (CS) and similarity constraints. The sparsity and self-similarity of the image blocks are taken as the constraints. The proposed method is summarized in the following steps. First, a dictionary classification method based on the measurement domain is presented. The image blocks are classified into smooth, texture and edge parts by analyzing their features in the measurement domain. Then, the corresponding dictionaries are trained using the classified image blocks. Equally important, in the reconstruction part, we use the CS reconstruction method to recover the HR brain MRI image, considering both nonlocal similarity and the sparsity of an image as the constraints. This method performs better both visually and quantitatively than some existing methods.

\end{abstract}
% Note that keywords are not normally used for peer review papers.

% For peer review papers, you can put extra information on the cover
% page as needed:
% \ifCLASSOPTIONpeerreview
% \begin{center} \bfseries EDICS Category: 3-BBND \end{center}
% \fi
%
% For peerreview papers, this IEEEtran command inserts a page break and
% creates the second title. It will be ignored for other modes.
\IEEEpeerreviewmaketitle

\section{Introduction}
% The very first letter is a 2 line initial drop letter followed
% by the rest of the first word in caps.
%
% form to use if the first word consists of a single letter:
% \IEEEPARstart{A}{demo} file is ....
%
% form to use if you need the single drop letter followed by
% normal text (unknown if ever used by the IEEE):
% \IEEEPARstart{A}{}demo file is ....
%
% Some journals put the first two words in caps:
% \IEEEPARstart{T}{his demo} file is ....
%
% Here we have the typical use of a "T" for an initial drop letter
% and "HIS" in caps to complete the first word.
\IEEEPARstart{O}{ver} the past decade, the brain Magnetic Resonance Imaging (MRI) has become one of the most important methods to diagnose the ailing brains. High-resolution (HR) images with sufficient details have found significant applications in medical imaging. Therefore, the requirement for image quality is fast increasing. However, due to the limitations of the physical resolution of the terminal devices or the bandwidth in the transmission process, it is difficult to obtain the high-resolution brain MR images that satisfy the basic requirement for applications. Attempts to resolve this dilemma have resulted in the development of an emerging research topic in image signal processing, known as super-resolution (SR) image reconstruction, which has been extensively studied in recent years. SR is an inverse problem that tackles the recovery of a high-resolution image from a single image or multiple low-resolution images of the same scene based on either specific a priori knowledge or reasonable assumptions about the imaging model that degrades the high-resolution image to the low-resolution ones.

SR image recovery is a terrible ill-posed problem because there are no sufficient low-resolution images, the blurring operators are unknown, and the solution from the recovery constraint is not unique. Many regularization methods have been presented to further improve the inversion of this underdetermined problem, such as \cite{jointmap,fastrobustsr,bayesiansr}. However, these reconstruction-based SR algorithms often lead to poor robustness and unsatisfied performance when the magnification factor is large. Thus, the reconstructed images may be overly smooth and absent of critical high-frequency details \cite{limitsofsr}. The interpolation-based SR approach is another type of SR method. Takeda et al. presented an interpolation algorithm based on the controllable kernel regression, which constructs the direction-controllable interpolation kernel function through a covariance matrix \cite{kernelsr}.  Li et al. applied different interpolation strategies for image blocks with various features. That is, in the bilinear interpolation for smooth regions and particular edge regions, the local covariance is used to adjust the interpolation coefficients \cite{newedgesr}. Recently, some structural adaptive interpolation methods have achieved good results. Yeon et al. proposed an edge-oriented local RBF interpolation algorithm \cite{nonlinearsr}. Romano et al. combined the interpolation with the nonlocal self-similarity and sparse representation of images and explored a new adaptive interpolation method \cite{singleimagesr}. However, high-resolution images recovered by these interpolation-based methods are prone to be overly smooth and have ringing and jagged artifacts.

Another category of SR methods is based on machine learning techniques, which seek to obtain the co-occurrence prior between low-resolution (LR) and high-resolution (HR) image patches. Freeman et al. first put forward using learning techniques to improve the image resolution. The authors used the Markov random field (MRF) to establish the corresponding relationship between the HR image block and the LR image block. The initial value of the HR image was obtained by interpolation. The lost high-frequency details of the HR image were recovered by learning and added with the initial value; then, the HR image was obtained \cite{examplebasedsr}. Sun et al. further improved this approach by applying the primal sketch priors to improve blurred edges, ridges, and corners. The SR methods using the convolutional neural network are presented in \cite{2018cnn1,2017cnn2}, which performed single- and multi-contrast super-resolution reconstructions simultaneously. Unfortunately, the aforementioned approaches generally require databases, which contain millions of HR and LR patch pairs and are therefore computationally intensive. In addition, there exist untrue high-frequency details, which are recovered by learning from external training databases.

The emergence of compressed sensing (CS) offers a new different perspective to address large underdetermined problems. CS can reconstruct sparse or compressible signals using fewer measurements than conventional methods without prior knowledge about the support of the signals. CS claims the inaccuracy of the conventional wisdom that the acquisition and reconstruction must follow Nyquist sampling theory \cite{candescs,donohocs,candesintrocs,candesstablesignal,candesnearoptimal,candesromberg}. This favorable and promising tool has proven to be applicable for various fields, including machine learning \cite{optics,sparsecs}, wireless communication \cite{wirelesscs1,wirelesscs2}, and medical imaging \cite{medicalcs1,medicalcs2}. Fortunately, due to its favorable property, CS can be applied to solve the SR problem. The application of CS and sparse representation in the field of SR recovery has captured the interest and attention of an enormous number of researchers in the past decade. The pioneer works can be traced to \cite{cssr2009,cssr2010yang,mrisr2012,srsr2012,singledlsr2012,srsr2016}. Sen et al. proposed a new algorithm to generate a super-resolution image from a single, low-resolution input without using a training data set \cite{cssr2009}. The CS theory was used to recover the HR image in magnetic resonance imaging \cite{mrisr2012}. Then, these methods were extended in \cite{srsr2012,singledlsr2012,srsr2016}. The authors presented new approaches to the single-image SR problem based on the sparse representation. In \cite{2013singledic}, Rueda et al. proposed a sparse-based super-resolution method coupling up high and low frequency information to reconstruct a high-resolution brain MR image. Several papers (e.g., \cite{du1,du2,du3,du4,du5}) have studied related sensing issues. However, these previous work failed to consider the combination of the sparse representation and nonlocal self-similarity.	Although much effort has been spent on improving the performance of SR recovery, an efficient and effective method has not been developed.

The purpose of this paper is to apply CS, the sparse representation and inherent similarity of an image to recover an HR image from a single LR image. It is of great interest and significance to address the questions in CS for ill-posed problems such as SR. In this paper, we have extended the previous work by paying attention to the nonlocal self-similarity of an LR image. We propose an image SR algorithm based on compressed sensing and self-similarity constraint. Because the difference of image blocks is not considered when training dictionaries, a dictionary classification method based on the measurement domain is proposed in the dictionary training part. Specifically, we use the linear relationship between images in the measurement domain and frequency domain to  classify the image blocks into smooth, texture and edge feature blocks in the measurement domain. The dictionaries for different blocks are trained by using different categories. Consequently, an LR image block of interest may be reconstructed using the most appropriate dictionary. If one merely learns the prior knowledge from the external image database, it tends to generate false details of the reconstructed HR image.

In our proposed method, we use the nonlocal similarity of the image to tentatively search for similar blocks in the whole image and present a joint reconstruction method based on CS and similarity constraints. The sparsity and self-similarity of the image blocks are used as the constraints. The proposed method is summarized in the following steps. First, a dictionary classification method based on the measurement domain is presented. The image blocks are classified into smooth, texture and edge parts by analyzing their features in the measurement domain. Then, the corresponding dictionaries are trained using the classified image blocks. Equally important, in the reconstruction part, we use the CS reconstruction method to recover the HR image considering both the nonlocal similarity and sparsity of an image as constraints. This approach results in visually and quantitatively better performance than some existing methods.

The remainder of this paper is organized as follows. In Sec.\ref{srbycs}, we briefly introduce the correlative theoretical basis, including CS, followed by the discussion of image SR using CS. Then, the proposed SR method based on CS and self-similarity is described in detail in Sec.~\ref{proposedmethod}. The explanation, illustration, and analysis of the experimental results are demonstrated in Sec.~\ref{results}. Finally, the summary of this paper is presented in Sec.~\ref{conclusion}.

\section{Image Super-Resolution Using CS}
\label{srbycs}
\subsection{Compressed Sensing}
\label{sec:cs}
 For completeness, we briefly introduce the fundamental background of CS.
 CS can reconstruct sparse or compressible signals using fewer measurements than the traditional approach uses. The advent of CS has tremendously affected signal acquisition and signal recovery \cite{candescs,donohocs,candesintrocs} because the compressibility or sparsity is of great significance. Suppose that $x$ is a discrete signal with size $n$; if it has no more than $r$ nonzero values, then $x$ is called ``$r$-sparse''. A signal may have no sparsity in some domains. Fortunately, we can always find a certain domain where signal $x$ can be considered sparse with an appropriate basis.

Considering the natural images, it is beneficial that there are sufficient bases and dictionaries so that the natural images in these bases become sparse or approximately sparse. A signal is considered ``approximately sparse'' if its amplitude exponentially decays. A signal is referred to as ``compressible'' if it has an approximately sparse representation on a certain basis.
Concerning a sparse signal, there is much less valuable ``information'' than unimportant data. CS can reconstruct sparse or compressible signals with much fewer samples than traditional methods.

Let $x$ ($x\in R^N$) be a discrete signal; $\theta$ represents its coefficients in a certain orthonormal basis \[y = \Phi x.\] Then, $x$ is $K$-sparse if only $K$ coefficients are nonzero. The procedure can be formulated as follows.
\[{\left\| {\mathbf{x}} \right\|_0}: = \left| {\left\{ {\ell :{x_\ell } \ne 0} \right\}} \right| = \#
 \left\{ {\ell :{x_\ell } \ne 0} \right\}
  \leq s. \]
 where ${\left\| {\mathbf{x}} \right\|_0}$ represents the $\ell_0$-norm of $\bf x$, which denotes the number of nonzero elements of $\bf x$. The $\ell_p$-norm is defined as
\[
	{\left\| {\mathbf{x}} \right\|_p} = {\left( {\sum\limits_{i = 1}^N {{{\left| {{x_i}} \right|}^p}} } \right)^{1/p}},\qquad 1\le p<\infty.
\]
We call a matrix ${\mathbf{\Phi }} \in {\mathbb{C}^{n \times N}}$ the measurement matrix; then, the recovery process is to reconstruct ${\bf x} \in{\mathbb C}^N$ from the measurements \[{\mathbf{y}} = {\mathbf{\Phi x}}. \]
If $n\ll N$, this problem is underdetermined and has no solution. Fortunately, CS theory finds that the solution can be obtained with extra information that ${\bf x}$ is $s$-sparse.

The original recovery method adopts $\ell_0$-minimization:
\begin{equation}
\begin{split}
	\label{eqn:01}%~\eqref{eq:l0minimizsCS}
	&\min \;\;\;{\kern 1pt} {\left\| {\mathbf{z}} \right\|_0}\quad \;\;\;{\kern 1pt} \\
&{\text{subject to }}\;\;\;{\kern 1pt} {\mathbf{\Phi z}} = {\mathbf{y}},
\end{split}
\end{equation}
but this is an NP-hard problem. Then, tractable substitutions are used, e.g., $\ell_1$-minimization:
\begin{equation}
\begin{split}
	\label{eqn:02}%~\eqref{eq:l1minimizsCS}
	&\min \;\;\;{\kern 1pt} {\left\| {\mathbf{z}} \right\|_1}\quad \;\;\;{\kern 1pt} \\
&{\text{subject to }}\;\;\;{\kern 1pt} {\mathbf{\Phi z}} = {\mathbf{y}},
\end{split}
\end{equation}
where \[
\begin{split}
&{\left\| {\mathbf{z}} \right\|_1} = \left| {{z_1}} \right| + \left| {{z_2}} \right| +  \cdots  + \left| {{z_N}} \right|\\
&{\text{for}}\quad \; {\mathbf{z}} = \left( {{z_1},{z_2},. . . ,{z_N}} \right) \in {\mathbb{C}^N}.
\end{split}
\]

Assuring the recovering ability of $x$ in Eq.(\ref{eqn:01}) via $\ell_1$-minimization and greedy algorithms is a sufficient condition to establish the RIP (restricted isometry property) of measurement matrix ${\mathbf{\Phi }}$: Given ${\mathbf{\Phi }} \in {\mathbb{C}^{n \times N}}$ and $s<N$, the RIC (restricted isometry constant) $\delta_s$ is defined as the smallest positive number such that
\begin{equation}
\label{eqn:03} %~\eqref{eq:definingRestrictedIsometryProperty}
\begin{split}
	&\left({1 - {\delta _s}} \right)\left\| {\mathbf{x}} \right\|_2^2 \leq \left\| {{\mathbf{\Phi x}}} \right\|_2^2 \leq \left( {1 + {\delta _s}} \right)\left\| {\mathbf{x}} \right\|_2^2\;\;\;{\kern 1pt} \\
&{\text{for all~}}{\mathbf{x}}\in{\mathbb C}^N\\
&\text{with~~}{\left\| {\mathbf{x}} \right\|_0\le s}.
\end{split}
\end{equation}
Eq. (\ref{eqn:03}) demands that at most $s$ columns of ${\mathbf{\Phi }}$ are well-conditioned. ${\mathbf{\Phi }}$ is said to satisfy the RIP with order $s$ when $\delta_s$ is small.

Many recovery methods are effective if the measurement matrix $\bf \Phi$ satisfies the RIP. More accurately, if the measurement matrix $\bf \Phi$ follows Eq. (\ref{eqn:03}) with
\begin{equation}
\label{eqn:04}  %~\eqref{pfander2011restricted:eq(4)}
	{\delta _{\kappa s}} < {\delta ^ \star }
\end{equation}
for appropriate constants $\kappa\ge 1$ and $\delta ^ \star$, then several algorithms can precisely reconstruct any $s$-sparse signals $\bf x$ from ${\mathbf{y}} = {\mathbf{\Phi x}}$. Furthermore, if $\bf x$ can be approximated by an $s$ sparse vector, then for noisy observations,
\begin{equation}
\label{eqn:05}
	{\mathbf{y}} = {\mathbf{\Phi x}} + {\mathbf{e}},\quad \;\;\;{\kern 1pt} {\left\| {\mathbf{e}} \right\|_2} \leq \alpha ,
\end{equation}
these algorithms can acquire the recovery ${{\mathbf{\tilde x}}}$ that satisfy an error bound as
\begin{equation}
\label{eqn:06}%~\eqref{pfander2011restricted:eq(5)}
	{\left\| {{\mathbf{x}} - {\mathbf{\tilde x}}} \right\|_2} \leq {C_1}\frac{1}{{\sqrt s }}{\sigma _s}{\left( {\mathbf{x}} \right)_1} + {C_2}\alpha ,
\end{equation}
where \[{\sigma _s}{\left( {\mathbf{x}} \right)_1} = \mathop {\inf }\limits_{{{\left\| {\mathbf{z}} \right\|}_0} \leq s} {\left\| {{\mathbf{x}} - {\mathbf{z}}} \right\|_1}\] represents the error of the best $s$-term approximation in $\ell_1$, and $C_1,C_2>0$ are constants.

\subsection{Super-Resolution based on Compressed Sensing}
\label{srbasecs}
The CS theory aims at solving the underdetermined problems and reconstructing a high-dimensional signal from fewer measurements than the traditional approach. For the SR problem, its goal is to recover a high-resolution image from a low-resolution one in the same scene. These two problems share a high similarity, so CS theory may be applied to solve the SR reconstruction problem. An SR problem may be viewed as the recovery process in the CS frame, where $Y$ can be considered the low-resolution image acquired as a measurement of the original high-resolution image $X$. Generally, matrix $M$, which degrades the HR image to an LR image in the SR problem, is considered the projection matrix in CS theory. The sparse basis is taken from the overcomplete dictionary $D$. In this work we consider only the case of a single image. Then, the process of solving the SR problem using CS theory is as follows:
\begin{equation}\label{equ29}
\begin{array}{l}
\alpha {\rm{ = argmin}}{\left\| \alpha  \right\|_0}\\
s.t.{Y_k} = {M_k}X = {M_k}D\alpha
\end{array}
\end{equation}
However, many factors must be considered, including the estimation of the degradation matrix, the method of training overcomplete dictionary $D$, and the specific reconstruction algorithm. The essence of applying the CS theory to SR is to make full use of the sparsity and fully excavate the intrinsic structural features of an image. SR based on CS theory has also made significant progress in recent years. The feasibility of applying CS theory to single-image SR has been proven in \cite{candesnearoptimal}. The mapping relationship between the HR dictionary and the LR dictionary has been established in \cite{cssr2010yang}.

This paper mainly studies how to reconstruct the HR image by using the sparsity of an image and the nonlocal similarity information inside the image. The principle of the SR algorithm based on the sparse representation is to regularize the image sparsity as a priori information. LR images are degraded, while the degradation model of HR to LR images is uncertain. The algorithm assumes that HR and LR images have similar geometric structures. Their sparse representations are approximate under a certain transform basis or redundant dictionary. We ensure the corresponding relationship between LR dictionary $D_l$ and HR dictionary $D_h$ atoms while training the dictionaries. Then, the relationship obtained by learning is applied to the current input image so that an HR image is reconstructed. This algorithm mainly includes the dictionary training process and reconstruction process, which are introduced in detail in Sec.~\ref{proposedmethod}.

\section{Proposed Method}
\label{proposedmethod}

This paper presents an image SR method based on the CS and nonlocal similarity. Because the difference of image blocks is not considered when training dictionaries, a dictionary classification method based on the measurement domain is proposed in the dictionary training part. Specifically, we use the linear relationship between images in the measurement domain and frequency domain to classify image blocks into smooth, texture and edge feature blocks in the measurement domain. The dictionaries for different blocks are trained using different categories. Consequently, an LR image block of interest may be reconstructed using the most appropriate dictionary. If one merely learns the prior knowledge from the external image database, it tends to generate untrue details of the reconstructed HR image. In our proposed method, we use the nonlocal similarity of the image itself to tentatively search for similar blocks in the whole image and present a joint reconstruction method based on CS and similarity constraints. The sparsity and self-similarity of the image blocks are taken as the constraints.

\subsection{Classified Dictionary Training}
\label{dictionary}

\begin{figure*}[hbpt]
 \centering
\includegraphics[width=18cm]{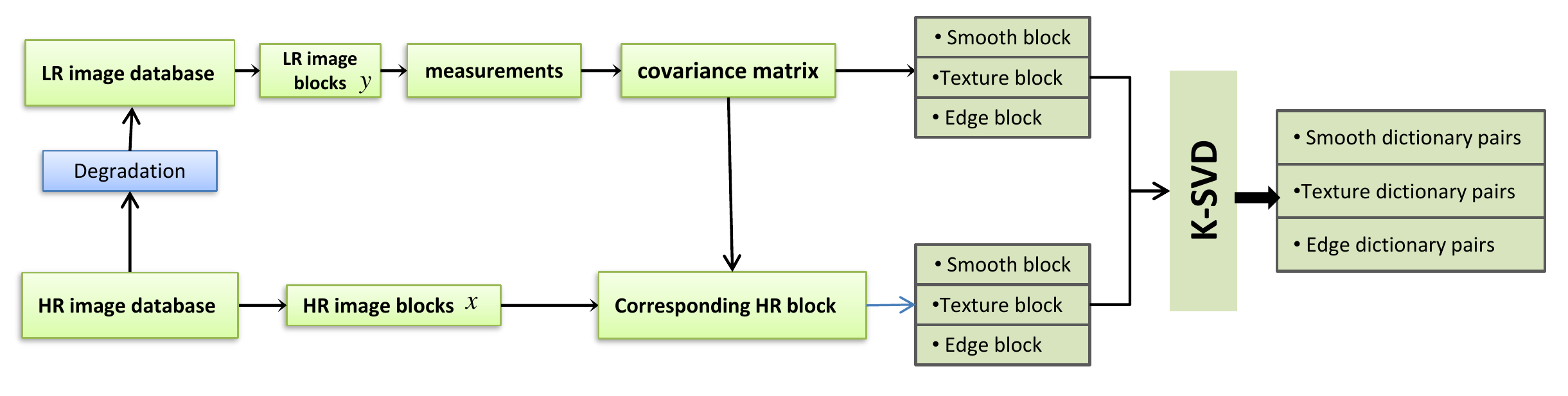}
\caption{Classified dictionary training based on the measurement domain.}
\label{fig:classifieddic}
\end{figure*}

The existing SR methods based on the sparse representation failed to consider the differences among sample blocks in the training dictionary. Remarkable differences between the input LR image and the sample database may lead to the poor quality of the reconstructed HR image. To overcome this problem, we propose a dictionary classification method based on the measurement domain. In our past work \cite{dicgenerationliu, diclearnliu,adaptiveadmmliu,liyinghua,liyinghua2}, we have proposed an adaptive ADMM algorithm with support and a maximum-likelihood dictionary to improve the ability of the dictionary to represent the signal sparsely. First, we classify the images in the sample database in the measurement domain; then, we use them to train different categories of dictionaries and reconstruct the input image block using the closest dictionary to improve the definition of the HR image. In our previous work \cite{guojie,motionestimationguojie}, we theoretically proved the approximately linear relationship between the cross-covariance matrixes in the measurement domain and frequency domain, which can be formulated as follows:
\begin{equation}
\label{liner}
{C_y} \approx \frac{n}{m}{C_q},
\end{equation}
where $m$ and $n$ represent the sample numbers in the measurement domain and frequency domain, respectively.
The images in the frequency domain and pixel domain are also closely related. Generally, an edge texture block is more sparse than a smooth block. We propose a classification method in the measurement domain using covariance matrixes to classify the image blocks in the training set. Different types of dictionaries are trained using different kinds of image blocks. The overall block diagram of the dictionary classification method based on the measurement domain is shown in Fig. \ref{fig:classifieddic}.

\begin{figure*}[hbpt]
 \centering
\includegraphics[width=17cm]{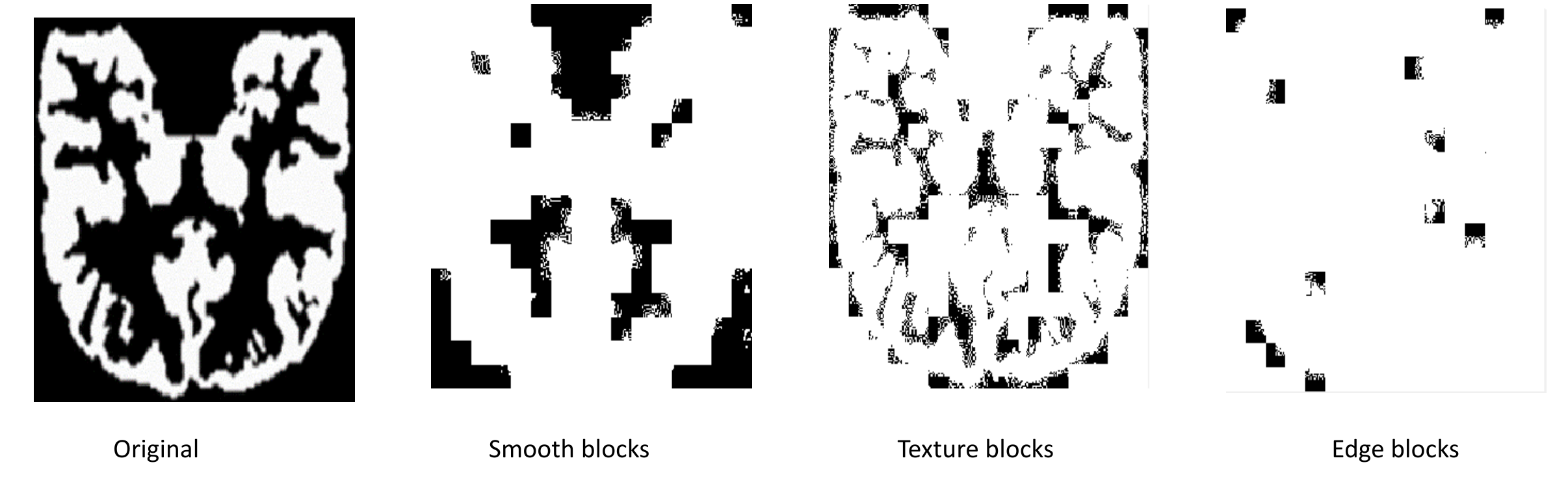}
\caption{Block classification result of a brain MRI image.}
\label{fig:blocks1}
\end{figure*}

We select the brain tissue MRI image as sample for the experiment to show the performance of classifying image blocks in the measurements. The images are divided into $8\times8$ blocks using the Gaussian matrix as the measurement matrix. Here, we use $T_1=3\times10^6$ and $T_2=3\times10^7$. The result is shown in Fig. \ref{fig:blocks1}.

The experimental result shows that the proposed classification method in this paper can better classify the image blocks into smooth blocks, texture blocks, and edge blocks. The sampling rate determines the amount of data to be sorted and processed. The lower the sampling rate, the fewer data there are to calculate. However, when the sampling rate is extraordinarily low, the measured value vector will be reduced accordingly. This process fails to contain all information of the original image, which leads to a large deviation in the classification results. The experimental experience value shows that if the sampling rate is not less than 0.4, better results can be guaranteed.

\subsection{Nonlocal Similarity of an Image}
\label{nonlocalsimi}
Natural images should preferably be rich in content and have certain repeatability in structural features. The repetitive information of an image has been widely used in image recovery, image denoising, and other issues. The fundamental principle of a nonlocal algorithm is to give different weight coefficients to the similar points of the current pixel using their linear combination to represent the current pixel. Therefore, the internal structure of the pixels can be maintained. Of course, the value of the coefficients dramatically depends on the similarity of the two pixels. The local phase theory holds that the similarity points of pixels exist in their adjacent local regions and that the neighborhood points have a high degree of approximation with the current point. However, the nonlocal similarity theory considers the repeatability of the image structure and holds that two pixels may have a higher degree of approximation even in the case of a longer spatial distance. Inspired by the nonlocal features of the image, this paper applies it to the SR algorithm to improve the quality of the HR image reconstruction.

\begin{figure}[htbp]
  % Requires \usepackage{graphicx}
  \includegraphics[width=2in]{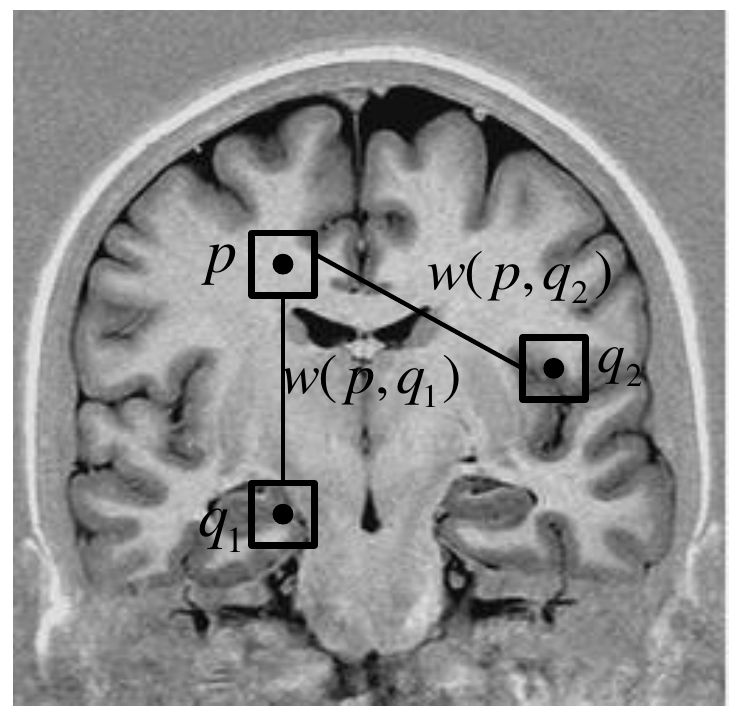}\\
  \caption{The similarity of pixels within an image. }
  \label{fig:similarityofimage1}
\end{figure}

Suppose that an image $I=\{I(i,j)\}|(i,j\in\Omega)$ has definition in $\Omega\subset N^2$, we use the linear combination of other similar pixel points with different weight coefficients to represent the current pixel $(i_0, j_0)$; its weighting value is:
\[NL(I)({i_0},{j_0}) = \sum\limits_{(i,j) \in I} {w(i,j)} I(i,j)\]
where the value of ${w(i,j)}_{(i_0, j_0)}$ is determined by the approximation degree of $(i,j)$ and $(i_0, j_0)$, which obeys $0\leq w(i,j)\leq1$ and $\Sigma{w(i,j)}=1$. Taking Fig. \ref{fig:similarityofimage1} as an example, $q_1$ is similar to $p$ in terms of gray value, whereas $q_2$ is significantly different from $p$. Therefore, the value of $w(q_1)$ is far greater than that of $w(q_1)$.

We define the pixel-centered window as the subset of $\Omega$: $N={\{N_{i,j}\}_{(i,j)\in\Omega}}$. We define the similarity between two central pixels by comparing the similarity of two window regions. Thus, the weight coefficient is proportional to the similarity of the two window areas and can be computed as:
\[w(i,j) = \frac{1}{{Z(i,j)}}\exp ( - \left\| {z({N_{{i_0},{j_0}}}) - z({N_{i,j}})} \right\|_2^2/{h^2})\]
where $Z(i,j) = \sum\limits_{i,j} {\exp ( - \left\| {z({N_{{i_0},{j_0}}}) - z({N_{i,j}})} \right\|_2^2/{h^2})} $ is the normalization factor, and $h$ is the decline rate of function.

\subsection{CS and Nonlocal Similarity-based Reconstruction}
The previous work has shown that the image blocks may frequently appear to be more similar in the interior of the image than in the exterior training database \cite{blocksimilar}.

Compared with the learning of the exterior library, more useful information can be obtained from the relevant information extracted from the interior of the image. However, for some image blocks, the information learned by themselves is limited and is not sufficient to reconstruct high-quality HR image blocks. Therefore, it is also necessary to obtain prior information through external learning to guide the current image block reconstruction. In this paper, we combine the nonlocal self-similar information of the image with the external dictionary and propose an SR method based on CS and self-similarity.

There are many similar blocks in the image and among different scales. A larger search area yields more similar blocks. To obtain more information contained within a single image, a tentative nonlocal search strategy is proposed in this paper. The adjacent regions of the current image block are helically squared matching to find similar blocks; for remote blocks, variable step-size searching is used according to the effect of similar blocks that have been found. This approach which may fully mine the similar information in the image and can be quickly completed.

\begin{figure*}[hbpt]
\centering
\includegraphics[width=16cm]{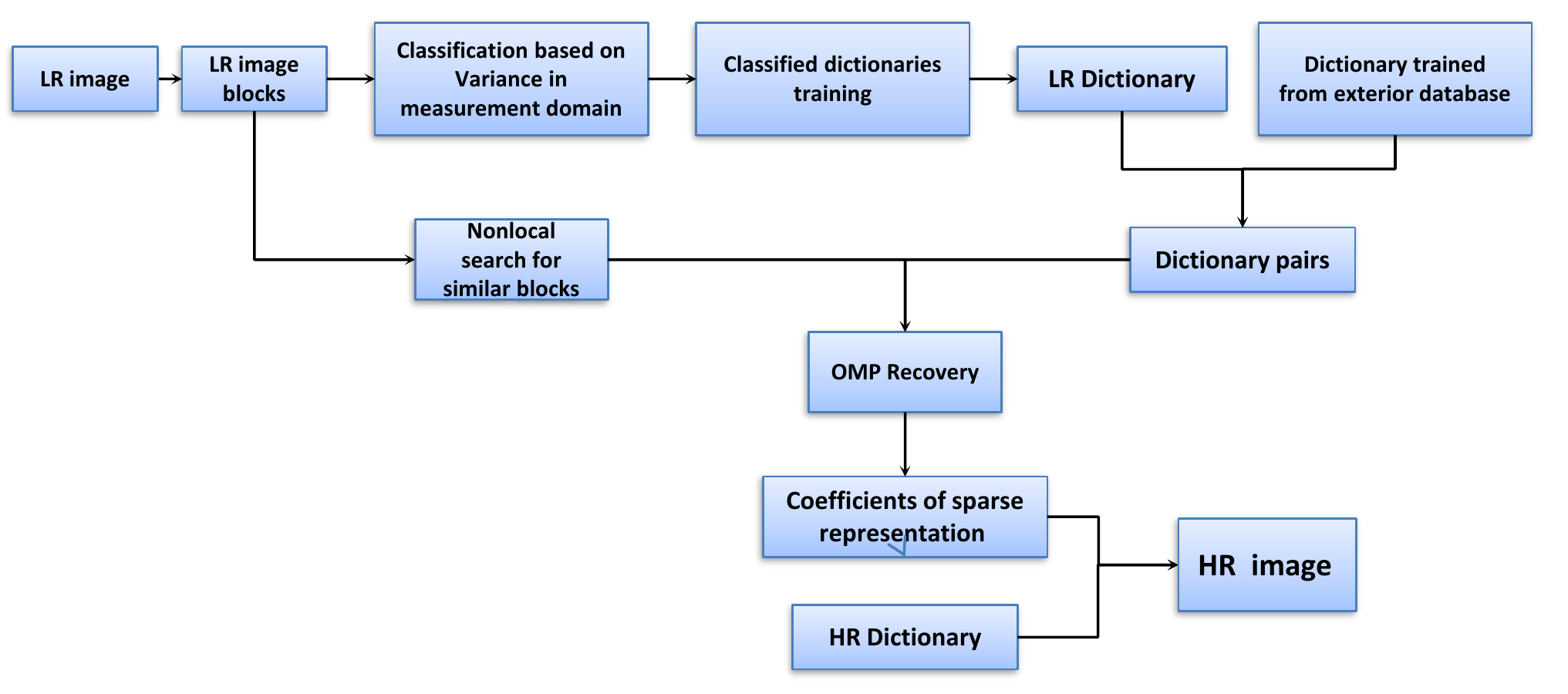}
\caption{Diagram of the reconstruction process in the proposed method. }
\label{reconstructionprocess}
\end{figure*}
The reconstruction process in this paper is shown in Fig. \ref{reconstructionprocess}. For any image block $y$ of an input LR image, a dictionary pair $(D_h, D_l)$ of the corresponding category is selected according to its variance. All of its similar blocks $S=\{y^1, y^2,..., y^n\}$ are found in the whole image. We add the self-similarity as the constraint, which requires coefficient $\alpha$ to be of high sparsity, and the HR image block represented by it has high similarity with its similar block $S$. The joint solution process using $s$ and $(D_h, D_l)$ can be expressed as:
\begin{equation}\label{equ314}
\begin{split}
\mathop {\mathop {\min }\limits_{\alpha ,{\alpha ^i}} }\limits_{i = 1,2,..n} \left\| {Fy - F{D_l}\alpha } \right\|_2^2 + \sum\limits_{{y^i} \in S} {\left\| {F{y^i} - F{D_l}{\alpha ^i}} \right\|_2^2}  \\
+ \lambda ({\left\| \alpha  \right\|_1} + \sum\limits_{i = 1}^n {{{\left\| {{\alpha ^i}} \right\|}_1}} ) + \sum\limits_{i = 1}^n {{\gamma _i}\left\| {{D_h}\alpha  - {D_h}{\alpha ^i}} \right\|_2^2}
\end{split}
\end{equation}

where $\alpha$ is the sparsity degree of current image block $y$, and ${\alpha}^i$ is the representation coefficient of $y^i$ on $D_l$. The first two items in the equation are used to guarantee the fidelity of the input LR image blocks, the two middle $l_1$ regularization items guarantee the sparsity of representation of the LR blocks on $D_l$, and the last item ensures the degree of approximation between the recovered HR image block and the similar block. The degree of approximation is controlled by ${\gamma}_i$:

\begin{equation}\label{equ315}
{\gamma _i} = \frac{1}{Z}\exp \{  - \frac{{\left\| {y - {y^i}} \right\|_2^2}}{{{h^2}}}\}
\end{equation}
where $Z$ is the normalization parameter.

The second, fourth and fifth items in Equation \ref{equ314} represent the nonlocal similarity information of the image blocks.
We obtain coefficient $\alpha$ by solving Equation \ref{equ314}. Then, the HR image block can be obtained by
\begin{equation}\label{equ316}
 x={D_h}\alpha
\end{equation}

By processing all LR blocks according to these steps, we recover the HR image $X$. Then, the IBP algorithm is used to more consistently guide $X$ to adjust along the direction with the image degradation model so that the final reconstructed HR image is consistent with the input LR image based on the image degradation model.

\section{Experimental Results}
\label{results}
The experiments are performed on both synthetic and real brain MRI images with the magnification factors of 2 and 4. We compare the results with the existing work \cite{nonlinearsr, 2018cnn1}. Their methods are denoted as Bicubic, and BSRCNN, respectively, for convenience.

We adopt the synthetic brain MRI images selected from Brainweb dataset \cite{brainweb} \footnote{http://brainweb.bic.mni.mcgill.ca/brainweb/}, MRT dataset \footnote{https://www.mr-tip.com/}, and the real MRI data from MIDAS dataset\footnote{http://insight-journal.org/midas/collection/view/190} which acquired with a 3T GE scanner at Brigham and Women's Hospital in Boston, MA and contains 10 normal and 10 schizophrenic patients.

\subsection{Evaluation Criterion}
Generally, the performance of the SR algorithm is evaluated from the following two perspectives:
\begin{itemize}
  \item Subjective evaluation. This method is mainly based on the visual perception of the human eyes to evaluate the quality of the image. Because individuals have different perceptions of the same image, this evaluation method is more influenced by subjective factors, which leads to the existence of individual differences.
  \item Objective evaluation. Since the LR test image in the SR algorithm is usually simulated by the degradation model of the HR image, there exists an original HR image, which is compared with the reconstructed image. The objective evaluation method is to determine the similarity between the recovered image and the original image using a calculation method. In this paper, two important criteria to evaluate the objective quality of SR methods are the PSNR (peak signal-to-noise ratio) and SSIM (structural similarity image measurement).
\end{itemize}
\begin{equation}\label{equ233}
 MSE = \frac{{\sum\nolimits_{i = 1}^M {\sum\nolimits_{j = 1}^N {{{({X_{ij}} - {Y_{ij}})}^2}} } }}{{M \times N}}
\end{equation}
\begin{equation}\label{equ234}
PSNR = 10{\log _{10}}\frac{{255 \times 255}}{{MSE}}
\end{equation}
\begin{equation}\label{equ235}
SSIM(X,Y) = \frac{{(2{\mu _X}{\mu _Y} + {C_1})(2{\delta _{XY}} + {C_2})}}{{(\mu _X^2 + \mu _Y^2 + {C_1})\left( {\delta _X^2 + \delta _Y^2 + {C_2}} \right)}}
\end{equation}
where $X$ is the original HR image, $Y$ is the recovered HR image, and $M$ and $N$ represent the size of the image.

\subsection{Visual Results}
Because the human eye system is sensitive to the luminance component, we only focus on the luminance $Y$ channel in the SR reconstruction of color images. The values of the chroma $C_b$ and $R_c$ channels are directly obtained using Bicubic upsampling. In the experiments, the size of the image block is $5\times5$, the overlap part is 4 pixels, and the number of dictionary atoms is 512.
The HR images in the test sets are downsampled by using the fuzzy downsampling matrix, and the corresponding LR images are generated by simulating the image degradation model. We use the proposed method and other reference algorithms to perform $2\times$ and $4\times$ SR reconstructions, respectively.

\begin{figure*}[hbpt]
\centering
\includegraphics[width=17cm]{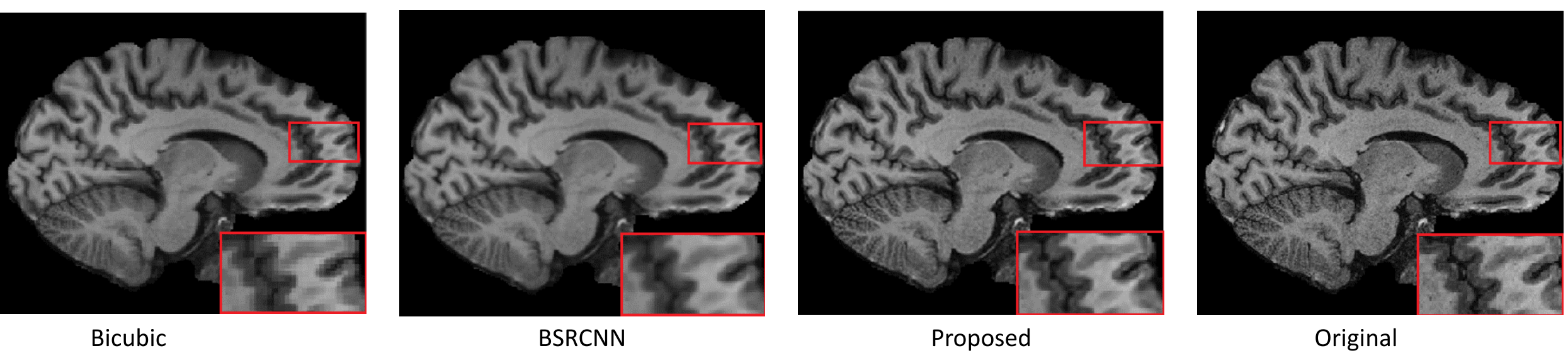}
\caption{Comparison of visual results with upscaling factor 2 (brain MRI image from Brainweb). }
\label{result1}
\end{figure*}

\begin{figure*}[hbpt]
\centering
\includegraphics[width=17cm]{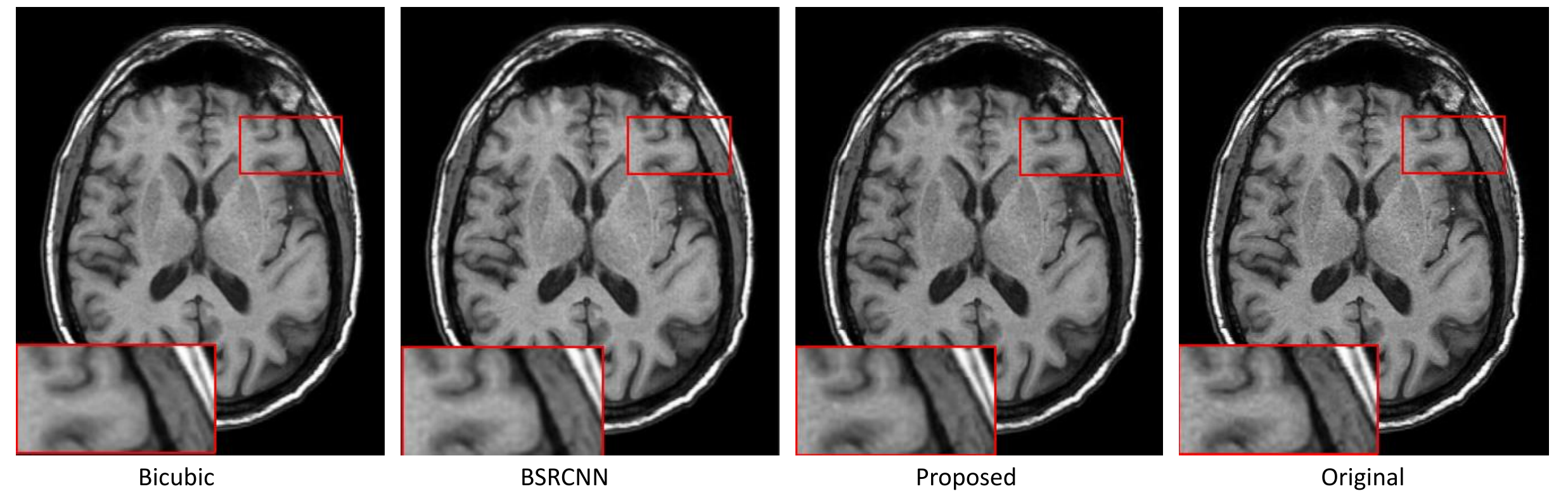}
\caption{Comparison of visual results with upscaling factor 2 (brain MRI image from MRT). }
\label{result2}
\end{figure*}

\begin{figure*}[hbpt]
\centering
\includegraphics[width=17cm]{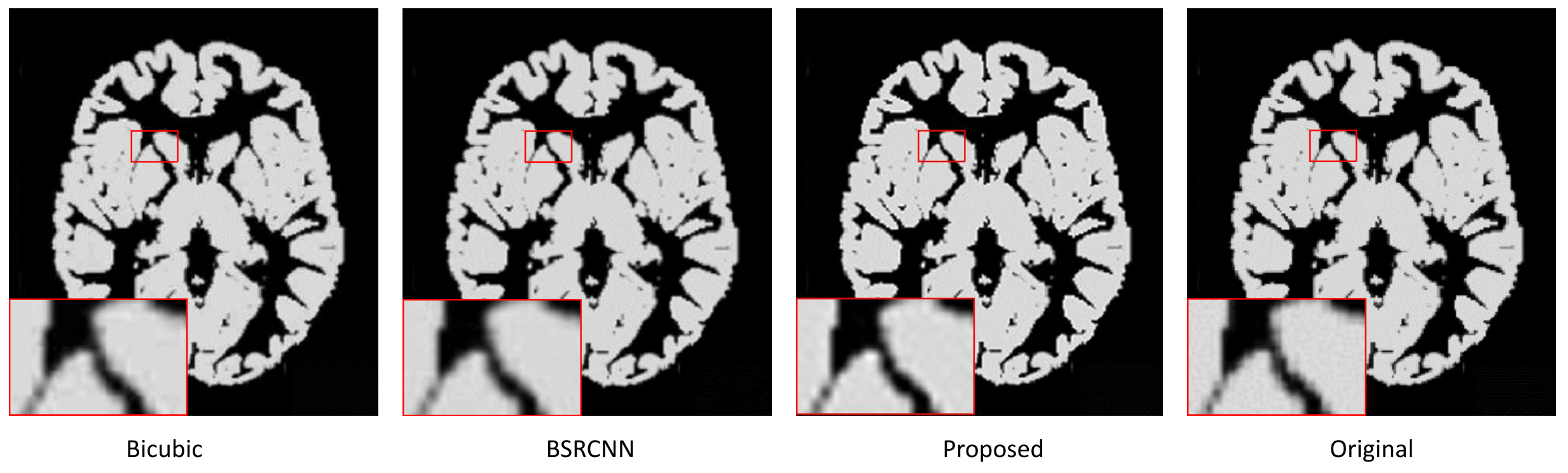}
\caption{Comparison of visual results with upscaling factor 4 (brain MRI image from Brainweb). }
\label{result3}
\end{figure*}

\begin{figure*}[hbpt]
\centering
\includegraphics[width=17cm]{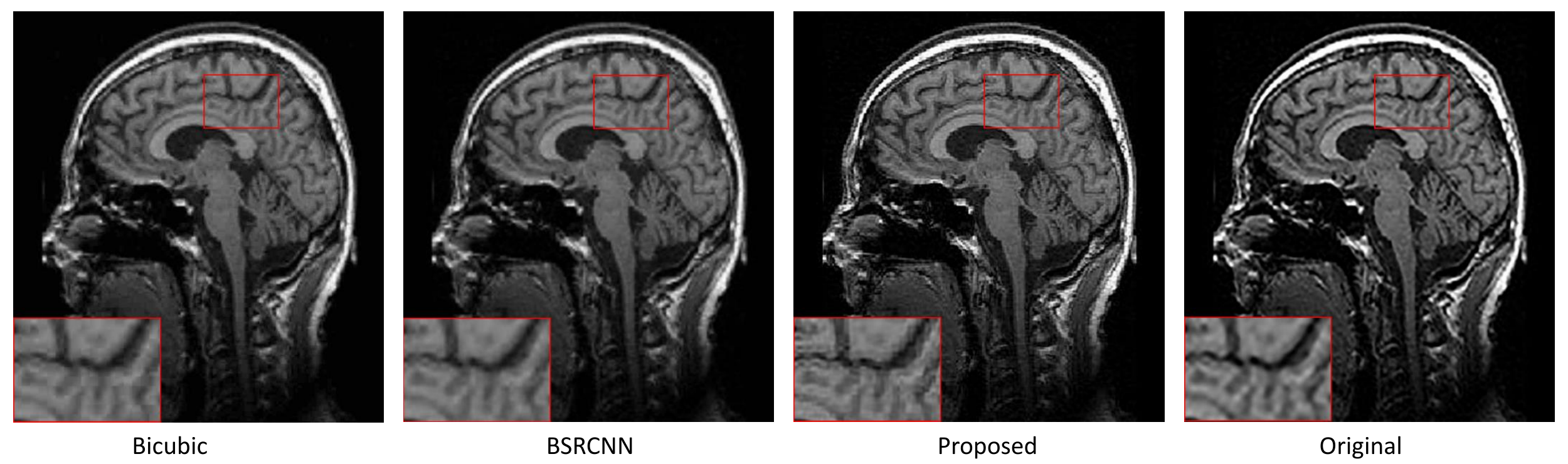}
\caption{Comparison of visual results with upscaling factor 4 (brain MRI image from MIDAS).}
\label{result4}
\end{figure*}

The visual results obtained using Bicubic, BSRCNN and the presented method are illustrated in Figs. \ref{result1}, \ref{result2}, \ref{result3}, and \ref{result4}.

Therefore, we conclude that the reconstructed images using our proposed method are rich in texture areas, have more natural outlines, and have no apparent zigzag effect.

\subsection{Objective evaluation}
In terms of objective quality, our proposal is compared with Bicubic \cite{nonlinearsr}, SROD \cite{2013singledic} and BSRCNN \cite{2018cnn1}. The performance is measured regarding PSNR and SSIM. We average the results of ten test images as the PSNR/SSIM value shown in the following tables. The results in Tabs. \ref{psnrssim1}, \ref{psnrssim2} and \ref{psnrssim3} show that the proposed method has better objective quality than other algorithms. Both PSNR and SSIM are improved: the PSNR value is increased by approximately 0.9-5.9dB, and the SSIM value is increased by approximately 0.02-0.14. Compared with the result using the magnification factor of 2, the improvement of 4 times magnification is more remarkable. Thus, when the magnification factor increases, we can obtain more significant improvement in HR image quality.

\begin{table}[htbp]
  \centering
  \caption{PSNR and SSIM of the recovered HR images with upscaling factor 2. }
    % Table generated by Excel2LaTeX from sheet 'Sheet1'
    \begin{tabular}{c|c|c|c|c|c}
    \toprule
          &       & Bicubic & SROD  & BSRCNN & Proposed \\
    \midrule
    \multirow{2}[4]{*}{Brainweb} & PSNR  & 21.51 & 28.56 & 32.99 & 35.36 \\
\cmidrule{2-6}          & SSIM  & 0.827 & 0.912 & 0.923 & 0.961 \\
    \midrule
    \multirow{2}[4]{*}{MRT} & PSNR  & 29.18 & 30.12 & 34.21 & 35.1 \\
\cmidrule{2-6}          & SSIM  & 0.872 & 0.91  & 0.924 & 0.941 \\
    \midrule
    \multirow{2}[4]{*}{MIDAS} & PSNR  & 27.69 & 30.23 & 31.1  & 33.21 \\
\cmidrule{2-6}          & SSIM  & 0.801 & 0.909 & 0.914 & 0.956 \\
    \bottomrule
    \end{tabular}%
  \label{psnrssim1}%
\end{table}%

\begin{table}[htbp]
  \centering
  \caption{PSNR and SSIM of the recovered HR images with upscaling factor 3.}
    % Table generated by Excel2LaTeX from sheet 'Sheet1'
    \begin{tabular}{c|c|c|c|c|c}
    \toprule
          &       & Bicubic & SROD  & BSRCNN & Proposed \\
    \midrule
    \multirow{2}[4]{*}{Brainweb} & PSNR  & 18.12 & 22.12 & 24.21 & 25.47 \\
\cmidrule{2-6}          & SSIM  & 0.712 & 0.745 & 0.785 & 0.842 \\
    \midrule
    \multirow{2}[4]{*}{MRT} & PSNR  & 21.34 & 24.61 & 27.49 & 29.31 \\
\cmidrule{2-6}          & SSIM  & 0.671 & 0.841 & 0.891 & 0.905 \\
    \midrule
    \multirow{2}[4]{*}{MIDAS} & PSNR  & 21.41 & 23.48 & 26.75 & 29.12 \\
\cmidrule{2-6}          & SSIM  & 0.715 & 0.756 & 0.814 & 0.816 \\
    \bottomrule
    \end{tabular}%
  \label{psnrssim2}%
\end{table}%

\begin{table}[htbp]
  \centering
  \caption{PSNR and SSIM of the recovered HR images with upscaling factor 4.}
    % Table generated by Excel2LaTeX from sheet 'Sheet1'
    \begin{tabular}{c|c|c|c|c|c}
    \toprule
          &       & Bicubic & SROD  & BSRCNN & Proposed \\
    \midrule
    \multirow{2}[4]{*}{Brainweb} & PSNR  & 16.39 & 20.41 & 21.31 & 23.15 \\
\cmidrule{2-6}          & SSIM  & 0.541 & 0.645 & 0.778 & 0.801 \\
    \midrule
    \multirow{2}[4]{*}{MRT} & PSNR  & 20.12 & 23.41 & 26.12 & 27.1 \\
\cmidrule{2-6}          & SSIM  & 0.563 & 0.674 & 0.741 & 0.756 \\
    \midrule
    \multirow{2}[4]{*}{MIDAS} & PSNR  & 19.54 & 21.41 & 24.56 & 26.45 \\
\cmidrule{2-6}          & SSIM  & 0.689 & 0.701 & 0.731 & 0.751 \\
    \bottomrule
    \end{tabular}%
  \label{psnrssim3}%
\end{table}%

\section{Conclusion}
\label{conclusion}
In this paper, we have extended the previous work by paying attention to the nonlocal self-similarity and the block classification of an LR image. We propose an image SR algorithm based on compressed sensing and self-similarity constraint. This proposed method is applied to solve the brain MRI super-resolution problem, and the satisfactory results may be acquired. Because the difference of image blocks are not considered when training dictionaries, a dictionary classification method based on the measurement domain is proposed in the dictionary training part. Specifically, we use the linear relationship between images in the measurement domain and frequency domain to classify the image blocks into smooth, texture and edge feature blocks in the measurement domain. The dictionaries for different blocks are trained using different categories. Consequently, an LR image block of interest may be reconstructed using the most appropriate dictionary. If one merely learns the prior knowledge from the external image database, it tends to generate untrue details of the reconstructed HR image. In our proposed method, we use the nonlocal similarity of the image to tentatively search for similar blocks in the whole image and present a joint reconstruction method based on the classified dictionaries and similarity constraints. The sparsity and self-similarity of the image blocks are taken as the constraints.

In summary, a dictionary classification method based on the measurement domain is presented. Then, the corresponding dictionaries are trained using the classified image blocks. Equally important, in the reconstruction part, we use the CS reconstruction method to recover the HR image, considering both nonlocal similarity and sparsity of an image as the constraints. This method visually and quantitatively performs better than some existing methods. To verify the performance of the proposed method, many experiments have been accomplished on both the synthetic and real brain MRI images. The experimental results indicate that the proposal enhances the quality of the recovered HR brain MRI image, and our method results in visually and quantitatively superior performance.

% use section* for acknowledgment

% Can use something like this to put references on a page
% by themselves when using endfloat and the captionsoff option.
\ifCLASSOPTIONcaptionsoff
  \newpage
\fi

\end{document}